%% file: cb-analysis.tex
\pdfoutput=1

\documentclass[11pt]{article}

\input{header}

\title{Analysis of Language Change in Collaborative Instruction Following}

\author{Anna Effenberger\textsuperscript{1},  Eva Yan\textsuperscript{2}\footnotemark[1], Rhia Singh\textsuperscript{2}\footnotemark[1], Alane Suhr\textsuperscript{1}, \and Yoav Artzi\textsuperscript{1} \\
  \textsuperscript{1}Cornell University \\
  \textsuperscript{2}City University of New York \\
  \texttt{ae347@cornell.edu} \hspace{10pt} \texttt{eyan0749@gmail.com} \\ \texttt{rhia.singh@macaulay.cuny.edu} \hspace{10pt} \texttt{\{suhr, yoav\}@cs.cornell.edu} }

\date{}

\begin{document}

\maketitle

\begin{abstract}
\input{00-abstract}

\end{abstract}

\renewcommand{\thefootnote}{\fnsymbol{footnote}}
\footnotetext[1]{Equal contribution.}
\renewcommand*{\thefootnote}{\arabic{footnote}}

\input{01-intro}

\input{02-scenario}

\input{03-analysis}

\input{04-discussion}

\input{05-ack}

\bibliography{main,local}
\bibliographystyle{acl_natbib}

\clearpage

\appendix

\input{a1-supp}

\end{document}

%% file: header.tex
\usepackage[]{emnlp2021}

\usepackage{times}
\usepackage{latexsym}

\usepackage[T1]{fontenc}

\usepackage[utf8]{inputenc}

\usepackage{microtype}

\usepackage{titlesec}
\titlespacing{\paragraph} {0pt}{1.0ex plus 0.1ex minus .1ex}{1em}

\usepackage{booktabs}
\newcommand{\ra}[1]{\renewcommand{\arraystretch}{#1}}

\definecolor{plotcolor1}{RGB}{35,127,215}
\definecolor{plotcolor2}{RGB}{46,159,52}
\definecolor{plotcolor3}{RGB}{227,118,18}

\newcommand*{\Courier}{\fontfamily{pcr}\selectfont}
\DeclareTextFontCommand{\textCourier}{\Courier}

\usepackage{tikz}
\usepackage{pgfplots}
\pgfplotsset{width=7cm,compat=1.3}
\usepgfplotslibrary{colorbrewer}

\usepackage{pgfplotstable}
\pgfplotstableset{col sep=comma}

\usepackage{balance}
\usepackage{xspace}

\newcommand{\xxcomment}[4]{\textcolor{#1}{[$^{\textsc{#2}}_{\textsc{#3}}$ #4]}}

\newcommand{\anna}[1]{\xxcomment{green}{A}{E}{#1}}
\newcommand{\eat}[1]{}

\newcommand{\cerealbar}{\textsc{CerealBar}\xspace}

\newcommand{\nlstring}[1]{{\em #1}}

%% file: 00-abstract.tex
We analyze language change over time in a collaborative, goal-oriented instructional task, where utility-maximizing participants form conventions and increase their expertise. 
Prior work studied such scenarios mostly in the context of reference games, and consistently found that language complexity is reduced along multiple dimensions, such as utterance length, as conventions are formed.
In contrast, we find that, given the ability to increase instruction utility, instructors increase language complexity along these previously studied dimensions to better collaborate with increasingly skilled instruction followers.

%% file: 01-intro.tex
\section{Introduction}\label{sec:intro}

Community language change in situated collaborative task-oriented scenarios has been studied with focus on reference games~\cite{Krauss1964ChangesIR,Clark1986-CLARAA-2,hawkins2017convention,hawkins2020characterizing,Hawkins2020:partners-popluations}, where two participants coordinate using language to select to a single item from a set of available items. 
These studies found that utility-maximizing participants trade surface-form linguistic complexity with established norms, as the familiarity and expertise of the interaction partners increase. 
In practice, this emerges as a reduction in utterance length and vocabulary size. 

We study the generality of these observations by analyzing language change in a collaborative instructional task, where instructors can specify multiple goals within a single instruction to increase their utility. 
This option, not present in reference games, creates competing incentives: increasing utility by issuing more goals in a single instruction versus decreasing language effort by utilizing established norms (e.g., by shortening instructions). 

\begin{figure}[ht] 
    \footnotesize
    \centering
    \fbox{\begin{minipage}[]{0.95\linewidth}
            \textbf{Decile 1:} \nlstring{get the card in front}
    \end{minipage}}
    \fbox{\begin{minipage}[]{0.95\linewidth}
        \textbf{Decile 5:} \nlstring{Collect the green square card in front of you.}
    \end{minipage}}
    \fbox{\begin{minipage}[]{0.95\linewidth}
    \textbf{Decile 10:} \nlstring{turn around on the trail, go straight and get 2 green circles, continue straight on the trail to the right side of the glacier and get 1 black triangle.}
    \centering
    \includegraphics[width=\textwidth,trim=70 320 422 157,clip=true]{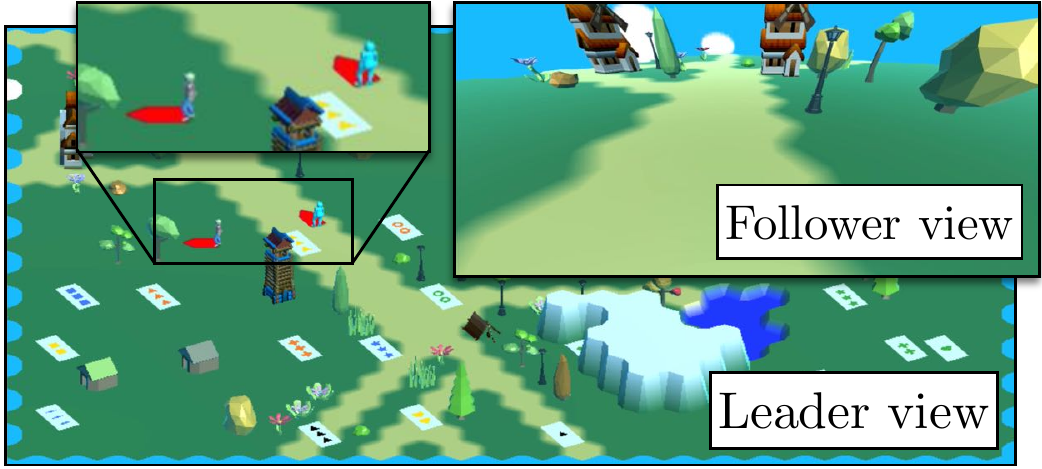}
    \end{minipage}}
    \caption{Leader instructions  in \cerealbar from games played at the beginning (Decile 1), middle (Decile 5), and end (Decile 10) of the community life. The differences between the instructions illustrate the linguistic change observed in the data. 
    The instruction from Decile 10 is paired with a snapshot from the game as the follower begins to execute it. 
    The leader (left) and follower (right) are highlighted in the center-left of the leader's view of the game, and the top right shows the follower's first-person view of the environment.}\label{fig:cb}
\end{figure}

We use the \cerealbar game environment and its accompanying dataset~\cite[Figure~\ref{fig:cb};][]{Suhr2019:cerealbar}. 
\cerealbar is a two-player, collaborative language game where players work together to collect sets of matching cards.
A leader plans which cards to include in the next set, and 
writes instructions to a follower describing tasks to accomplish.
In contrast to reference games~\cite{Krauss1964ChangesIR}, the language in \cerealbar is primarily instructional rather than referential, and the game allows players to complete a dynamic number of tasks per instruction and game.

Similar to previous studies, we observe language change over time along the same dimensions. 
But, unlike in reference games, we observe utterance-level linguistic complexity increases. %
Our study illustrates that the formation of common ground among interaction participants does not necessarily reduce language complexity, and may even come with an increase in complexity.
Understanding how humans use language to collaborate in settings with flexible \eat{levels of} utility is key to building natural language systems that effectively collaborate with \eat{human} users over time. 
Our analysis code can be found at \href{https://github.com/lil-lab/CB-analysis}{github.com/lil-lab/CB-analysis}.

%% file: 02-scenario.tex
\section{Scenario and Data Overview}\label{sec:scenario}

We use the \cerealbar game and accompanying dataset~\cite{Suhr2019:cerealbar} in our analysis. 
\cerealbar is a collaborative, two-player game, where a leader and a follower collect matching sets of cards by moving in an environment. 
The game is turn-based, and each player has a limited number of steps per turn. 
The leader both collects cards and instructs the follower using natural language.\footnote{All utterances are in English.} 
The follower executes leader instructions. 
The players' abilities differ: the leader observes the complete environment and plans sets to collect; the follower only observes what is ahead, but has more steps per turn. 
For each set made, players receive one point and additional turns, allowing them to complete more sets. 
Success requires the players to collaborate via natural language: the leader must write informative instructions to the follower, and the follower must efficiently follow these instructions.
Figure~\ref{fig:cb} shows a snapshot of the game.

The \cerealbar dataset contains 1{,}202 human-human game interactions collected \eat{using Amazon Mechanical Turk }over the course of four months.
Workers were randomly assigned as leader or follower for each interaction. 
The collection process created a Wizard-of-Oz setup: the system user, as the leader, provides instructions and acts in the world, and the human follower is a wizard, executing instructions to emulate the desired system behavior. 
We only use interactions from the training split for our analysis. 
We prune interactions by inexperienced workers, as classified when the data was  collected, to focus on the impact of experience.\footnote{Appendix~\ref{sec:supp:data} describes this pruning process.} 
In total, we consider 795 interactions.
Table~\ref{tab:datastats} provides basic statistics of the data we use. 
\citet{Suhr2019:cerealbar} used these data to train models, while we study how the language changes. %

\begin{table}[t]
    \centering
    \ra{0.95}
    \setlength{\tabcolsep}{2pt}
    \footnotesize
    \begin{tabular}{@{}l c c c@{}}
    \toprule
     & Mean & Median & Max \\
    \midrule
    Interaction Score (\# Card Sets) & 8.8 & 10.0 & 19 \\
    \# Instructions / Interaction & 22.0 & 26.0 & 41 \\
    \# Tokens / Instruction & 14.4& 13.0 & 55 \\
    \midrule
    Vocabulary Size & \multicolumn{3}{c}{3{,}499}\\
    Total \# Instructions & \multicolumn{3}{c}{17{,}524} \\
    \bottomrule 
    \end{tabular}
    \caption{Statistics of analyzed data.
    } \label{tab:datastats}
\end{table}

%% file: 03-analysis.tex
\section{Data Analysis}\label{sec:analysis}

\begin{figure}[t]
\footnotesize
\centering
\begin{tikzpicture}
\pgfplotsset{
    xtick={1,2,3,4,5,6,7,8,9,10},
    xmin=0, xmax=11, 
    width=4.5cm, height=3cm
}
\begin{axis}[
  ymin=0, ymax=50,
  ybar, ticks=none, bar width = 3,
]
\addplot[ybar, fill=plotcolor3]
  coordinates{
    (1,33)
    (2,40)
    (3,34)
    (4,47)
    (5,39)
    (6,43)
    (7,36)
    (8,40)
    (9,41)
    (10,41)
}; 
\end{axis}

\begin{axis}
[
  ymin=0, ymax=50,
]
\addplot[mark=*,plotcolor1]
  coordinates{
    (1, 33)
    (2,17)
    (3,8)
    (4,13)
    (5,6)
    (6,1)
    (7,1)
    (8,3)
    (9,0)
    (10,0)
};
\end{axis}
\end{tikzpicture}
\begin{tikzpicture}
\pgfplotsset{
    xtick={1,2,3,4,5,6,7,8,9,10},
    xmin=0, xmax=11,
    width=4.5cm, height=3cm
}
\begin{axis}
[
  ymin=0, ymax=15,
]
\addplot[mark=square*,plotcolor2]
  coordinates{
  (1,3.76)
  (2,5.46)
  (3,7)
  (4,7.69)
  (5,9.33)
  (6,9.12)
  (7,10.2)
  (8,11.1)
  (9,11.8)
  (10,12.3)
};
\end{axis}
\end{tikzpicture} 
\mbox{Community Lifetime Decile}
    \caption{Community size (left) and mean game score (right) over deciles of community lifetime. On the left, the bars show total active players and the curve shows only the number of new players that joined per decile. }\label{fig:gamestats}
    \vspace{+3pt}
\end{figure}
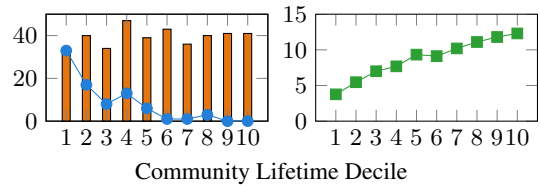

To analyze trends over the data collection period, we split the data chronologically into 10 deciles of roughly equal size (79 or 80 interactions).
An average of 40 workers participated in each decile (Figure~\ref{fig:gamestats}, left). 
The community stabilized after Decile 4, as worker recruitment slowed and the community was split by expertise.\footnote{Appendix~\ref{sec:supp:deciles} provides decile details.}

Interaction goals are increasingly achieved over time.
Mean score per game increases from 3.8 to 12.3 ($p < 0.0001$) (Figure~\ref{fig:gamestats}, right).\footnote{We use a two-sided t test at $\alpha = 0.05$ for all calculations of significance when comparing means.} 
Execution efficiency and game expertise also improve.\footnote{ Appendix~\ref{sec:supp:performance} details this improvement.}
Our focus is how leader language - the sole communication conduit - changes to enable these gains.

We design our analysis to be as similar as possible to existing work on reference games~\cite{hawkins2020characterizing}, which shows that certain language aspects are simplified as community conventions form. 
\cerealbar allows for a different realization of common ground development than previously studied reference games, and we observe trends that are in contrast to this line of prior work.

\begin{figure}[t]
  \footnotesize
  \centering
  \begin{tikzpicture}
  \pgfplotsset{
      xtick={1,2,3,4,5,6,7,8,9,10},
      xmin=0, xmax=11,legend style={nodes={scale=0.75, transform shape}}, legend pos=south east, width=6cm, height=4cm
  }
  \begin{axis}
  [
    axis y line*=left,
    ymin=700, ymax=1400,
    xlabel=Community Lifetime Decile,
    ylabel=Vocabulary
  ]
  \addplot[mark=square*,plotcolor1]
    coordinates{
    (1,753)
    (2,743)
    (3,911)
    (4,965)
    (5,992)
    (6,1067)
    (7,988)
    (8,1084)
    (9,1142)
    (10,1070)}; \label{Vocabulary}
  \end{axis}
  
  \begin{axis}[
    axis y line*=right,
    axis x line=none,
    ymin=10, ymax=16,
    ylabel=Utterance Length
  ]
  \addlegendimage{/pgfplots/refstyle=Vocabulary}
  \addlegendentry{Vocabulary}
  \addplot[mark=*,plotcolor3]
    coordinates{
    
    (1,11.9)
    (2,13.8)
    (3,14.3)
    (4,14.1)
    (5,14.3)
    (6,14.9)
    (7,14.5)
    (8,14.4)
    (9,14.1)
    (10,14.1)}; \addlegendentry{Utterance length}

  \end{axis}
  \end{tikzpicture} 
      \caption{Vocabulary and utterance length over deciles.}\label{fig:lenvocab}
      \vspace{+3pt}
  \end{figure}
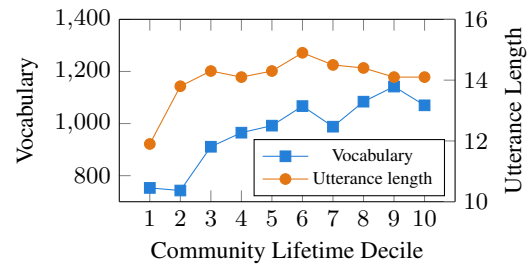

\paragraph{Instruction Length and Vocabulary}
Mean\footnote{All means over instructions are first computed within each game, then across games. This weighs all games equally, rather than upweighing longer, higher-scoring games.} instruction length increases from 11.9 to 14.1 tokens\footnote{We use NLTK for tokenization, lowercase all tokens, and use the {\tt autocorrect} library for typo correction.} ($p < 0.0001$) over time, while vocabulary size increases from 752 to 1{,}070 unique tokens (Figure~\ref{fig:lenvocab}). 
This contrasts with reference games, where utterance length and vocabulary size reduce~\cite{Clark1986-CLARAA-2,hawkins2017convention}. 
Some of the words added more specifically describe props or movements.
However, the overall trend is relatively complex, and identifying clear patterns 
 likely requires a more targeted scenario design.%

\paragraph{Syntactic Complexity}

We analyze syntactic trends using parts-of-speech (POS) tags and dependency trees.\footnote{We use spaCy~\cite{spacy2} for POS tagging and dependency parsing.} 
We do not observe a significant difference in usage of closed- and open-class POS tags, as seen in reference games~\cite{hawkins2017convention}. 
We observe change in the relative use of verbs, nouns, conjunctions, determiners, and numerals.\footnote{Appendix~\ref{sec:supp:syntax} provides details.} 
Notably, the proportion of conjunctions of all tokens increases from 0.060 to 0.067 ($p = 0.0026$).\footnote{We use a one-sided $z$ test at $\alpha = 0.05$ for calculations of significance when comparing proportions.}
The proportion of instructions that contain a conjunction also increases from 0.0495 to 0.0707 ($p = 0.0113$). 
Qualitatively, this accompanies an increased use of ordered sentential conjunctions, often to specify multiple tasks in a single utterance (e.g., \nlstring{once you get that card, turn around and \eat{then} go left and get the 1 green circle card}).

\begin{figure}[t]
  \footnotesize
  \centering
  \begin{tikzpicture}\pgfplotsset{
      xtick={1,2,3,4,5,6,7,8,9,10}, xmin=0, xmax=11,
  legend pos=south east, legend style={nodes={scale=0.75, transform shape}}, width=6cm, height=4.75cm
  }
  \begin{axis}
  [
    axis y line*=left,
    ymin=11.5, ymax=15.25,
    xlabel=Community Lifetime Decile,
    ylabel=Length,
  ]
  \addplot[mark=x,gray]
    coordinates{
    (1,11.9)
    (2,13.8)
    (3,14.3)
    (4,14.1)
    (5,14.3)
    (6,14.9)
    (7,14.5)
    (8,14.4)
    (9,14.1)
    (10,14.1)
  }; \label{Utterance Length}
  \end{axis}

  \begin{axis}[
    axis y line*=right,
    axis x line=none,
    ymin=.85, ymax=1.1,
    ylabel=Normalized Complexity
  ]
  \addlegendimage{/pgfplots/refstyle=Utterance Length}
  \addlegendentry{Average utterance length}
  \addplot[mark=square*,plotcolor1]
    coordinates{
      (1,0.934)
      (2,0.976)
      (3,0.989)
      (4,1.02)
      (5,0.997)
      (6,0.992)
      (7,0.998)
      (8,0.985)
      (9,0.974)
      (10,1.00)
  }; \addlegendentry{Branch factor}

  \addplot[mark=*,plotcolor2]
    coordinates{
      (1,1)
      (2,1.03)
      (3,1)
      (4,1.01)
      (5,1)
      (6,.997)
      (7,1.01)
      (8,.993)
      (9,.993)
      (10,.988)
  }; \addlegendentry{Maximum depth}
  
  \addplot[mark=square*,plotcolor3]
    coordinates{
      (1,0.941)
      (2,0.963)
      (3,0.996)
      (4,0.998)
      (5,1.01)
      (6,1.00)
      (7,1.00)
      (8,0.980)
      (9,0.973)
      (10,0.987)
  }; \addlegendentry{Maximum width}
  
  \end{axis}
  \end{tikzpicture} 
  \caption{Average syntactic branching factor, maximum depth, and maximum width across deciles. 
  We also plot the mean utterance length for reference.  }\label{fig:syntax}
  \vspace{+3pt}
\end{figure}
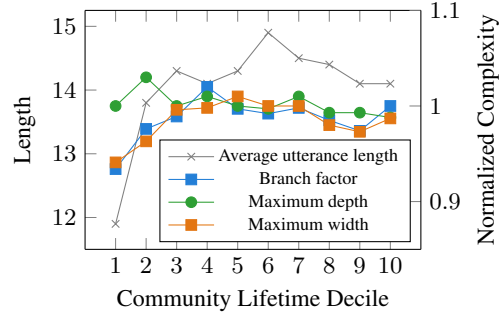

We compute three measures of syntactic complexity using dependency trees~\cite{xu-reitter-2016-convergence}: (a) maximum depth: the longest path from root to a leaf; (b) maximum width: the maximum out-degree of any node; and (c) average branching factor: the average out-degree of non-leaf nodes.\footnote{We further explain the syntactic measures and provide example instructions for illustration in Appendix~\ref{sec:supp:syntax}.}  
We normalize all measures to control for utterance length. 
Figure~\ref{fig:syntax} shows these statistics over time.
Maximum width and branching factor increased from 0.941 to 0.987 ($p = 0.0483$) and from 0.934 to 1.00  ($p = 0.0051$), indicating increased descriptiveness.
Maximum depth did not significantly change, indicating embedded clause use proportional to length, as expected when increasingly combining instructions with conjunctions.

\begin{figure}[t]
\footnotesize
  \centering
  \begin{tikzpicture}\pgfplotsset{
      xtick={1,2,3,4,5,6,7,8,9,10},
      xmin=0, xmax=11,
      width=4.5cm, height=3cm
  }
  \begin{axis}
  [
    ymin=0.9, ymax=1.1,
    ytick distance=0.1,
  ]
  \addplot[mark=square*,plotcolor1]
    coordinates{
      (1,0.998)
      (2,1.06)
      (3,1.04)
      (4,1.06)
      (5,1.03)
      (6,1.03)
      (7,1.04)
      (8,1.00)
      (9,0.983)
      (10,1.03)
  }; %
  \addplot[mark=*,plotcolor3]
    coordinates{
      (1,0.916)
      (2,0.953)
      (3,0.965)
      (4,0.990)
      (5,0.986)
      (6,0.975)
      (7,0.972)
      (8,0.973)
      (9,0.967)
      (10,0.983)
  }; %
  \end{axis}
  \end{tikzpicture} 
  \begin{tikzpicture}
  \pgfplotsset{
      xtick={1,2,3,4,5,6,7,8,9,10},
      xmin=0, xmax=11,
      width=4.5cm, height=3cm
  }
  \begin{axis}
  [
    ymin=0.9, ymax=1.1,
    yticklabels={,,}
  ]
  \addplot[mark=square*,plotcolor1]
    coordinates{
      (1,1.00)
      (2,1.06)
      (3,1.04)
      (4,1.04)
      (5,1.03)
      (6,1.0)
      (7,1.03)
      (8,1.00)
      (9,0.984)
      (10,1.02)
  }; %
  \addplot[mark=*,plotcolor3]
    coordinates{
      (1,0.923)
      (2,0.937)
      (3,0.976)
      (4,0.974)
      (5,0.999)
      (6,1.00)
      (7,0.985)
      (8,0.962)
      (9,0.965)
      (10,0.969)
  }; %
  \end{axis}
\end{tikzpicture} 
\mbox{Community Lifetime Decile}
      \caption{Average dependency branching factor (left) and maximum width (right) over deciles split to games that were above (blue) / below (orange) that decile's median game score. }\label{fig:branchoverscore}
      \vspace{+5pt}
\end{figure}
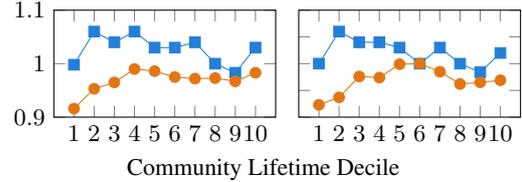

We observe similar trends when measuring these statistics  when comparing low- and high-scoring games (Figure~\ref{fig:branchoverscore}).
Higher scoring games had, on average, instructions with significantly higher width and branching factor. 
In Decile 1, language in games scoring 1 point and 16 points had an average normalized branch factor of 0.915 and 1.02. 
However, games in the lower 50\% of scores showed a higher increase in syntactic complexity over time.

Overall, our syntactic analysis shows an increase in language complexity is required to describe more tasks within a single instruction. We do not observe a gradual drop of redundant modifiers and descriptors~\cite{hawkins2017convention}. This may be because potential referents do not pose as much ambiguity as the abstract shapes often used in reference games~\cite{Clark1986-CLARAA-2}. 

\paragraph{Changes in References}
We see no significant development of niche idioms, in contrast to reference games with abstract shapes \cite{hawkins2020characterizing}. 
This is likely due to concreteness and familiarity of the referents in \cerealbar, allowing players to rely on common background knowledge with little ambiguity. 
We observe change in the relative frequency of references to specific objects over time. \eat{(Figure \ref{fig:Landmarkprop})}
We consider seven object classes: building, road, foliage, rock, ice, water, and light.\footnote{Appendix~\ref{sec:supp:refclass} describes this classification process.}
The proportion of instructions containing a reference to ice, light, and buildings increase from 0.006 to 0.022 ($p = 0.0006$), from 0.015 to 0.027 ($p = 0.0188$), and from 0.056 to 0.073 ($p = 0.0436$).
The ratios of other references are stable. 
Leaders likely choose references to balance informativity and effort. 
Foliage objects are common and require more effort to differentiate, while buildings and ice clearly vary.
Lights, though common, were often referred to with other objects to clarify location.

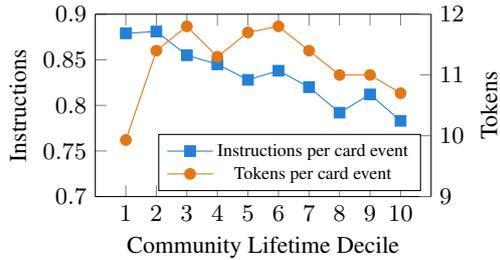
\begin{figure}
\footnotesize
\centering
\begin{tikzpicture}
\pgfplotsset{
    xtick={1,2,3,4,5,6,7,8,9,10},
    xmin=0, xmax=11,
legend pos=south east, legend style={nodes={scale=0.75, transform shape}}, width=6cm, height=4cm
}
\begin{axis}
[
  axis y line*=left,
  ymin=0.7, ymax=.9,
  xlabel=Community Lifetime Decile,
  ylabel=Instructions
]
\addplot[mark=square*,plotcolor1]
  coordinates{
    (1,0.879)
    (2,0.881)
    (3,0.855)
    (4,0.845)
    (5,0.828)
    (6,0.838)
    (7,0.820)
    (8,0.792)
    (9,0.812)
    (10,0.783)
  
}; \label{Instructions}
\end{axis}

\begin{axis}[
  axis y line*=right,
  axis x line=none,
  ymin=9, ymax=12,
  ylabel=Tokens
]
\addlegendimage{/pgfplots/refstyle=Instructions}
\addlegendentry{Instructions per card event}
\addplot[mark=*,plotcolor3]
  coordinates{
    (1, 9.93)
  (2,11.4)
  (3,11.8)
  (4,11.3)
  (5,11.7)
  (6,11.8)
  (7,11.4)
  (8,11.0)
  (9,11.0)
  (10,10.7)
}; \addlegendentry{Tokens per card event}

\end{axis}

\end{tikzpicture}

    \caption{The number of instructions and tokens required for a card event over deciles. 
    Analysis considers only instructions marked complete by the follower. }\label{fig:infostats}
\end{figure}

\paragraph{Language Effort} %
Leaders in \cerealbar mainly instruct followers to complete card events to ultimately select valid card sets. 
We measure language effort with respect to this objective as the number of tokens and instructions per card event (Figure~\ref{fig:infostats}). 
This notion of effort is similar to utterance cost in speaker-listener pragmatic models~\cite{Goodman2016PragmaticLI}.
The number of instructions per card event decreases from 0.879 to 0.783  ($p = 0.0102$), indicating leaders effectively pack more tasks into fewer instructions -- often multiple card events into one instruction in later deciles (Figure~\ref{fig:cb}). 
This change correlates with structural changes. 
For example, conjunctions are useful to pack more tasks into single instructions; the correlation across deciles between the proportion of instructions containing a conjunction and the number of instructions per card event is $r = -0.8243$. 
The high negative correlation indicates that the change in conjunction use aligns with the increase in goals (i.e., cards to select) packed per instruction. 
The number of tokens per card event initially increases from 9.9 to 11.8, then decreases to 10.7.
This may be because, initially, followers require more verbose instructions and leaders experiment with the level of description, but as conventions form, this verbosity is less needed to understand instructions.

The reduction in the number of tokens per goal later on corresponds to the reduction in utterance length observed in reference games~\cite{hawkins2017convention}, although it is manifested differently as the overall surface-form is not simplified (i.e., via shorter utterances), unlike in reference games.   
Given the opportunity to increase utility, leaders choose to take advantage of followers' increased expertise and efficiency by using more complex language to pack more goals into each instruction.

%% file: 04-discussion.tex
\vspace{-3pt}
\section{Discussion and Related Work}\label{sec:disc}
\vspace{-3pt}

The \cerealbar scenario is related to reference games~\cite{Krauss1964ChangesIR,Clark1986-CLARAA-2,hawkins2017convention,monroe2017colors,He:17dialogue,Udagawa:19,Haber:19}, 
which require two players to agree on a single referent from a set via dialogue. 
\cerealbar differs in several ways. 
It allows only unidirectional language communication, %
and utterances in \cerealbar are instructions specifying desired follower behavior with any number of tasks to complete (i.e., with flexible utility), not a description of a single target referent. 
\eat{The incentives also differ: the players' goal is to collect as many valid sets as quickly as possible.}

These differences lead to different language dynamics.
In reference games, \citet{hawkins2020characterizing} observed the development of specialized reference phrases for ambiguous shapes, which allows players to reduce their utterances' length and syntactic complexity.
Given that \cerealbar objects are generally unambiguous and familiar, players do not begin with overly verbose references, and have less potential for reduction to more concise references.
In contrast, we observe increased instruction length and complexity. 
Leaders issue an increasing number of tasks to the follower per instruction, utilizing the flexibility afforded by \cerealbar's design.
This less constrained scenario better reflects real-life collaborations, where participants complete many tasks to achieve complex goals.

Our observations show the competing effects of cost-minimization and utility-maximization. 
The formation of common ground and expectations on partners' behavior enables leaders to use language differently to convey more information-dense instructions to optimize game performance. 
This is aligned with the expectation of better communication grounding between community members in \citet{Clark1981:definite-mutual-knowledge}, and with how grounding in \citet{Clark1986-CLARAA-2} manifests as reduced complexity when utterance utility is fixed. 
Because there are conflicting forces at work in \cerealbar, common ground is realized differently.

The most related setup to \cerealbar is the Cards task~\cite{Djalali12:cards-preference,Potts:12}, where two players collect a single set of cards. 
It uses four static environments and studies dialogue, not instructions.
\citet{Djalali11:cards-qud} showed Cards players increase the interaction complexity by developing a rich common ground, including terms for the fixed board locations.
This is less likely with the randomly generated \cerealbar environments. 
Utterances in Cards also become shorter, potentially due to the predefined number of goals.%

Language complexity also increases in communities where users jointly build a natural-language-like programming language~\cite{wang-etal-2017-naturalizing,gavran2018precise}.
This scenario differs from ours in lacking explicit collaboration on tasks, focusing on a learned programming language rather than natural language, and training a single model, differently from our many-listeners community.

The language dynamics observed in \cerealbar contrast with those previously observed in reference games, providing evidence that gradual formation of common ground among interaction participants does not necessarily result in reduced complexity of sentences, and may even result in increased complexity. 
Our conclusions do not void nor mutually exclude previous work, but illustrate the complexity of language change over time in a community. 
An important direction for future work is controlled studies to observe the effects of scenario design on the interaction between the development of common ground and language change.

%% file: 05-ack.tex
\section*{Acknowledgments}

This research was supported by NSF under grants No. 1750499, 1750499-REU, and DGE-1650441. It also received support from a Google Focused Award, the Break Through Tech summer internship program, and a Facebook PhD Fellowship. 
We thank Chris Potts and Robert Hawkins for early discussions that initiated this analysis; and Ge Gao and Forrest Davis for their comments.

%% file: a1-supp.tex
\section{Reproducibility Checklist Details}

All computation was done on a personal laptop.
The \cerealbar data was acquired from \url{https://github.com/lil-lab/cerealbar}.
 
\section{Data Details}\label{sec:supp}

\subsection{Selection of Interactions for Analysis}\label{sec:supp:data}

The data we use was not collected specifically for this analysis, but during data collection for model development by \citet{Suhr2019:cerealbar}.
We use 795 of the 960 interactions in the original training split of the data for our analysis, pruning the rest to avoid games that include inexperienced players later in the community's life. 
This prevents the language of novice workers from affecting our analysis after the more experienced community had stabilized, which would potentially suppress convention formation trends observed in existing literature about reference games~\cite{hawkins2020characterizing}.
During the original data collection process, after 367 of the 960 total training interactions were collected, the community was split into junior and senior workers. 
Junior workers became senior upon gaining adequate experience.
A junior worker could request to be moved to the senior pool after they had played at least one game as a follower and at least one game as a leader where they earned at least one point with their partner, and they seemed to be following the game rules.
Workers who  performed well before the split were included in the senior pool.
We do not consider games from the junior pool.

\subsection{Decile Details}\label{sec:supp:deciles}

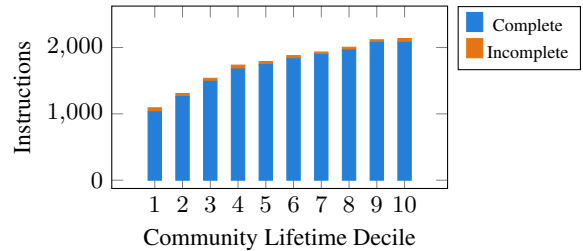
\begin{figure}[t]
  \footnotesize
  \centering
  \begin{tikzpicture}
  \pgfplotsset{
      xtick={1,2,3,4,5,6,7,8,9,10},
      xmin=0, xmax=11, width=6cm, height=4cm
  }
  \begin{axis}[
x tick label style={/pgf/number format/1000 sep=},
  ymin=0, ymax=2500, xlabel=Community Lifetime Decile, ylabel=Instructions, xtick align=inside,
  ybar stacked, yticklabel style={
        /pgf/number format/fixed,
        /pgf/number format/precision=5
},
scaled y ticks=false, bar width = 5,
	enlargelimits=0.05,
    legend pos=outer north east, legend style={nodes={scale=0.75, transform shape}},
]
  \addplot+[ybar, color=plotcolor1]
    plot coordinates{
      (1, 1031)
      (2,1261)
      (3,1490)
      (4,1676)
      (5,1743)
      (6,1830)
      (7,1895)
      (8,1962)
      (9,2079)
      (10, 2078)
  }; 
  \addplot+[ybar, color=plotcolor3] plot coordinates{
    (1,60)
    (2,46)
    (3,46)
    (4,59)
    (5,46)
    (6,48)
    (7,37)
    (8,42)
    (9,38)
    (10,57)
}; 
  \label{Instructions per decile}
  \legend{Complete, Incomplete}
  \end{axis}
  
  \end{tikzpicture}
  \caption{The number of instructions for each decile, distinguished by whether they were marked as complete by the follower.}\label{fig:numinstructions}
  \vspace{+7pt}
  \end{figure}

All deciles span a relatively short period of time except the sixth decile, which includes a pause in data collection (Table~\ref{tab:deciletimes}). 
The pause did not significantly effect community membership or performance. 
Figure~\ref{fig:numinstructions} shows the number of instructions per decile, distinguished by complete and incomplete instructions. 
Incomplete instructions occur at the end of an interaction, when there is insufficient time or turns to complete the instruction.  
Figure~\ref{fig:decile_interaction_length} shows mean interaction length in each decile. 
Figure~\ref{fig:decile_path_length} shows follower path lengths per instruction across each decile.

\begin{table*}[t]
\footnotesize
\centering
\begin{tabular}{@{}l c c c c@{}}
\toprule
Decile & Game IDs & Lower Time Limit  & Upper Time Limit & Time (Days) \\ 
\midrule
1      & 1-79     & 2019-01-27 20:05:00 UTC    & 2019-02-02 15:39:00 UTC         & 5.815278               \\
2      & 80-159   & 2019-02-02 15:39:00 UTC    & 2019-02-02 20:24:00 UTC         & 0.197917              \\
3      & 160-238  & 2019-02-02 20:24:00 UTC    & 2019-02-03 00:25:00 UTC         & 0.167361              \\
4      & 239-318  & 2019-02-03 00:25:00 UTC    & 2019-02-04 00:15:00 UTC         & 0.993055            \\
5      & 319-397  & 2019-02-04 00:15:00 UTC    & 2019-02-04 03:09:00 UTC         & 0.120833            \\
6      & 398-477  & 2019-02-04 03:09:00 UTC    & 2019-04-15 19:27:00 UTC         & 70.6375                \\
7      & 478-556  & 2019-04-15 19:27:00 UTC    & 2019-04-15 23:44:00 UTC         & 0.178472             \\
8      & 557-636  & 2019-04-15 23:44:00 UTC    & 2019-04-16 20:06:00 UTC         & 0.848611              \\
9      & 637-715  & 2019-04-16 20:06:00 UTC    & 2019-04-16 22:50:00 UTC         & 0.113889              \\
10     & 716-795  & 2019-04-16 22:50:00 UTC    & 2019-04-17 03:43:00 UTC & 0.203472             \\
\bottomrule
\end{tabular}
\caption{Time limits of the division into deciles. The last column is the total amount of time elapsed during a decile. All lower time limits are inclusive. All upper time limits are exclusive, except the last one, which is inclusive. }
\label{tab:deciletimes}
\end{table*}

\input{supp_boxplots}

\section{Additional Analysis Details}

\subsection{Interaction Performance}\label{sec:supp:performance}
Several measures demonstrate an increase in player expertise. We analyze interaction performance through how many moves are taken per each instruction, the occurrence of de-selection card events, and instruction queuing behavior. We find that followers become better at following instructions and leaders at creating efficient plans.

\paragraph{Optimal Path Length Deviations}\label{para:optimalpath}
We measure how leaders utilize the larger number of steps per turn available to followers through the length of the shortest possible path corresponding to each instruction. We compute this shortest path using the observed start and end positions of the human follower, ensuring that the path avoids obstacles and completes card events completed by the original follower.
The mean length of the shortest path per instruction increases over the community lifetime from 6.66 to 7.97 moves ($p < 0.0001$). This corresponds to the increase we observe in the number of goals described in each instruction, which likely requires more steps. 

Concurrently, we see improvements in follower instruction execution, measured through the excess moves taken by follower: the difference between the number of moves the follower took and the shortest possible path corresponding to each completed instruction. 
Over time, the number of excess steps compared to the shortest paths decreased from 3.67 to 2.36 moves ($p < 0.0001$).
Figure~\ref{fig:wandering} visualizes this increase in average optimal path length per instruction and decrease in moves taken in excess of this optimal path.
The reduction in excess moves is especially notable given the increase in the moves required per instruction, indicating the absolute decrease observed is due to an even higher decrease in the probability of follower errors.

\begin{figure}[t]
\footnotesize
\centering
\begin{tikzpicture}
\pgfplotsset{
    xtick={1,2,3,4,5,6,7,8,9,10},
    xmin=0, xmax=11,legend style={nodes={scale=0.75, transform shape}}, width=6cm, height=4cm
}
\begin{axis}
[
  axis y line*=left,
  ymin=2, ymax=4,
  xlabel=Community Lifetime Decile,
  ylabel=Excess Moves
]
\addplot[mark=square*,plotcolor1]
  coordinates{(1,3.67) (2,3.58) (3,3.10) (4,3.10) (5,2.77) (6,2.77) (7,3.10) (8,2.38) (9,2.30) (10,2.36)}; \label{Excess moves}
\end{axis}

\begin{axis}[
  axis y line*=right,
  axis x line=none,
  ymin=6.5, ymax=9.75,
  ylabel=Shortest Distance
]
\addlegendimage{/pgfplots/refstyle=Excess moves}
\addlegendentry{Excess moves}
\addplot[mark=*,plotcolor3]
  coordinates{(1,6.66)(2,7.45) (3,7.33) (4,7.57) (5,7.95) (6,7.59) (7,7.96) (8,8.19) (9,8.01) (10,7.97)}; \addlegendentry{Shortest distance}
\end{axis}
\end{tikzpicture} 
    \caption{Excess follower moves and shortest possible distance per leader instruction.}\label{fig:wandering}
\end{figure}

\paragraph{Card De-selections} \label{para:deselection}
We also study the occurrence of card de-selections, which often reflect error correction. 
In ideal gameplay, no de-selection events should be observed, as they require additional steps and only correct for a mistakenly selected card not to be part of the current target set. 
We observe that player errors decrease: the proportion of card events (the selection or de-selection of a single card) that are de-selections decreases from 7.86\% to 4.52\% ($p = 0.0018$). 
Figure~\ref{fig:deactivate} shows the percentage of card events initiated by either player that are de-selections.

\begin{figure}[t]
\footnotesize
\centering
\begin{tikzpicture}
\pgfplotsset{
    xtick={1,2,3,4,5,6,7,8,9,10},
    xmin=0, xmax=11,
    xlabel=Community Lifetime Decile,
    width=6cm, height=3cm
}
\begin{axis}[
  ymin=0.04, ymax=0.08, yticklabel style={
        /pgf/number format/fixed,
        /pgf/number format/precision=5
},
scaled y ticks=false]
\addplot[mark=*, color=plotcolor3] coordinates{
    (1,0.0786)
    (2,0.0591)
    (3,0.0569)
    (4,0.0717)
    (5,0.0582)
    (6,0.0706)
    (7,0.0513)
    (8,0.0504)
    (9,0.0602)
    (10,0.0452)
}; 
\end{axis}
\end{tikzpicture} 
    \caption{Proportion of all card events, initiated by both followers and leaders, that were de-selections.}\label{fig:deactivate}
\end{figure}

\paragraph{Instruction Queuing}\label{para:queuebehavior}
The \cerealbar setup allows a leader to plan ahead by queuing multiple instructions to the follower at a time. 
\eat{
For example, if the leader knows that the follower is able to complete a set early during the follower's next turn, they may queue additional instructions to instruct about the next set. 
To automatically align instructions to action sequences, the \cerealbar game only allows followers to see the instruction they are currently working on and previous instructions.
Marking an instruction as complete reveals the next instruction in the queue, if one was given. 
}
For example, to efficiently use all of the follower's moves, a leader may send two instructions: one which tells them to complete the set, and another that tells them to move towards a card which will make up the next set.
A larger queue indicates longer-term leader planning. 
Alternatively, the leader could include the additional information in one instruction without queuing more instructions. 
We analyze this queuing behavior as a potential alternative explanation: the leaders may improve how they relay information with better planning, rather than changing the content of their instructions.

\begin{figure}[t]
\footnotesize
\centering
\begin{tikzpicture}\pgfplotsset{
    xtick={1,2,3,4,5,6,7,8,9,10},
    xmin=0, xmax=11, ,legend pos=north east, legend style={nodes={scale=0.75, transform shape}}, width=6cm, height=5cm
}
\begin{axis}
[
  ymin=0, ymax=4, xlabel=Community Lifetime Decile, 
  ylabel=Instructions in Queue,
]
\addplot[mark=square*,plotcolor1]
  coordinates{
    (1,2.32)
    (2,2.44)
    (3,2.51)
    (4,2.74)
    (5,2.39)
    (6,2.49)
    (7,2.51)
    (8,2.31)
    (9,2.41)
    (10,2.38)
}; 
\addlegendentry{Max}
\label{Max}
\addplot[mark=triangle*,plotcolor3]
  coordinates{
    (1,1.43)
    (2,1.50)
    (3,1.46)
    (4,1.60)
    (5,1.47)
    (6,1.46)
    (7,1.48)
    (8,1.42)
    (9,1.45)
    (10,1.43)
}; \addlegendentry{Begin-turn}
\addplot[mark=*,plotcolor2]
  coordinates{ (1,0.694) (2,0.762) (3,0.678) (4,0.783) (5,0.675) (6,0.631) (7,0.642) (8,0.587) (9,0.594) (10,0.592)}; \addlegendentry{End-turn}
\end{axis}

\end{tikzpicture} 
    \caption{Instruction-queuing behavior over time.%
    }\label{fig:queue}
    \vspace{+7pt}
\end{figure}

\definecolor{purp}{RGB}{199,124,255}
\pgfplotscreateplotcyclelist{poscolorlist}{
{plotcolor1,mark=*}, %
{purp,mark=*}, %
{magenta,mark=x}, %
{blue,mark=x}, %
{cyan,mark=x}, %
{plotcolor3,mark=*}, %
{plotcolor2,mark=*}, %
{red,mark=x}, %
{violet,mark=*}%
}

\pgfplotstableread[col sep = comma]{posdecomp.csv}\posdata
 
 \begin{figure}[ht!]
 \footnotesize
\begin{tikzpicture}
\pgfplotsset{
yticklabel style={/pgf/number format/fixed,/pgf/number format/precision=5}, scaled ticks=false, %
xtick={1,2,3,4,5,6,7,8,9,10}, xmin=0, xmax=11, 
width=6cm, height=5cm}
\begin{axis}
[ymin=0.03, ymax=0.175,xlabel=Community Lifetime Decile,ylabel=Part of Speech Proportion,legend pos=outer north east, legend style={nodes={scale=0.75, transform shape}},cycle list name=poscolorlist] colormap/jet
\pgfplotsinvokeforeach{1,...,9}{ %
    \addplot table[x index = {0}, y index=#1]{\posdata};} %
\legend{Verb, Det, Prep, Adj, Adv, Conj, Num, Aux, Noun}
\end{axis}
\end{tikzpicture} 
    \caption{Ratio of language that is a specified part of speech over time. Parts of speech of particular interest are plotted with filled markers.\vspace{+7pt}}\label{fig:POSprop}
\end{figure}

\begin{figure}[ht!] 
    \footnotesize
    \centering
    \fbox{\begin{minipage}[]{0.95\linewidth}
            \textbf{Dep = 0.83, Wid = 0.93, Bch = 0.83} 
            
            \nlstring{ turn to the left to see one yellow sqaure} %
    \end{minipage}}
    \fbox{\begin{minipage}[]{0.95\linewidth}
            \textbf{Dep = 1.14, Wid = 1.03, Bch = 0.96} 
            
            \nlstring{ go forward one and to your left is orange}
    \end{minipage}}
    \fbox{\begin{minipage}[]{0.95\linewidth}
        \textbf{Dep = 1.58, Wid = 0.66, Bch = 0.65} 
        
        \nlstring{take the green card with 3 symbols in front of you}
    \end{minipage}}
    \fbox{\begin{minipage}[]{0.95\linewidth}
    \textbf{Dep = 0.79, Wid = 1.26, Bch = 1.01} 
    
    \nlstring{Head straight towards the blue plus card, but don't pick it up. Continue past it, on the left of it.}
    \end{minipage}}
    \caption{Selected instructions to illustrate the different measures of complexity, namely: maximum depth (dep), maximum width (wid), and average branching factor (bch). All measures normalized for length.}\label{fig:complexityex}
    \vspace{+5pt}
\end{figure}

We measure the size of the queue at the beginning and end of follower turns, and the maximum queue size reached during a game.
Figure~\ref{fig:queue} shows queue statistics over time. 
Begin-turn queue size directly measures how leaders plan via queuing instructions, as no instructions are queued during the follower's turn. 
Begin-turn and maximum queue size did not change significantly over time. 
This relative stability indicates that game play improvements were not primarily due to leaders planning ahead across separate instructions; rather, they can be attributed more to the changes of language within instructions. 
End-turn queue size sampling indicates the efficiency of player collaboration. 
From the first to last decile, the average end-turn queue size decreases from 0.694 to 0.592 instructions. 
This indicates that followers become more efficient over time, completing more instructions per turn. 
This aligns with our analysis of follower efficiency (Section~\ref{para:optimalpath} and Figure~\ref{fig:wandering}). 

\eat{
\pgfplotstableread[col sep = comma]{indivlife_depth.csv}\depthdata
\pgfplotstableread[col sep = comma]{indivlife_width.csv}\widthdata
\pgfplotstableread[col sep = comma]{indivlife_branchfct.csv}\branchdata
\begin{figure*}[ht!]
  \footnotesize
  \centering
  \begin{tikzpicture}\pgfplotsset{ xmin=0, xmax=45,width=14cm,height=5cm}
    \begin{axis}
    [ymin=0.9, ymax=1.2,xlabel=$n$-th Game Played as Leader,ylabel=Normalized Complexity,legend pos=north west, legend style={nodes={scale=0.75,transform shape}}]
    \addplot[mark=square*,plotcolor3] table [x index = {1}, y index = {3}] {\depthdata};
    \addlegendentry{Maximum depth}
    \addplot[mark=*,plotcolor2] table [x index = {1}, y index = {3}] {\widthdata};
    \addlegendentry{Maximum width}
    \addplot[mark=triangle*,plotcolor1] table [x index = {1}, y index = {3}] {\branchdata};
    \addlegendentry{Average branching factor}
    \end{axis}
  \end{tikzpicture} 
      \caption{Average branching factor, maximum width, and maximum depth over individual lifetime. Individual age is measured in absolute game count as leader. For each age, we average the statistics across all users that reach this age. The increase in variance for higher ages is because of the relatively small number of individuals reaching this age. Therefore, these measurements are less reliable.}\label{fig:IndivLifeCopmlexity}
\end{figure*}
}

\subsection{Syntactic Complexity}\label{sec:supp:syntax}

\paragraph{Part-of-Speech Analysis}\label{para:POS} 
To compute the ratio of POS use, we treat each decile of community life as a bag of words, dividing the total tag count of each POS by the total token count in each decile. In our analysis, we combine the spaCy tags $\langle$sconj$\rangle$ (subordinating conjunction) and $\langle$cconj$\rangle$ (coordinating conjunction) into one conjunction class, and the tags nouns and proper nouns into one noun class. Figure~\ref{fig:POSprop} shows the proportion of the nine most common POS tags used in \cerealbar instructions: verbs, determiners, prepositions, adjectives, adverbs, conjunctions, numerals, auxiliary verbs, and nouns.

\paragraph{Syntactic Complexity Analysis}\label{para:complexity_details} 
For each utterance, we measure the branching factor, maximum width, and maximum depth of its dependency parse. 
Dependency tree depth indicates how many embedded clauses the utterance has, whereas width-related measures indicate how many modifiers are stacked in one sub-tree. 
Intuitively, increased width-related metrics indicate more descriptive utterances, whereas increased depth indicates more compounded phrases.  
Figure~\ref{fig:complexityex} provide examples to illustrate these differences.

We normalize these measures by the utterance length following \citet{xu-reitter-2016-convergence}. 
Formally, let $X_n$ be the set of all utterances in our data with a length of $n$ tokens.
The average of metric $S$ (e.g., maximum width) across all utterances of length $n$ in our data is: 
\begin{equation}
\overline{S}(n) = \frac{1}{|X_n|} \sum_{x \in X_n} s(x)\;\;.
\end{equation}
For each utterance $x$ with length $n$, we compute the normalized measure for the utterance:
\begin{equation}
    s'(x) = \frac{s(x)}{\overline{S}(n)}\;\;.
\end{equation}

\eat{
\paragraph{Syntactic Complexity and Score}\label{para:complexity_score} 

We studied the relation between game score and syntactic trends. 
We divided games in each decile into two equally-sized sets of high and low-scoring games, and plotted the average dependency branching factor (Figure~\ref{fig:branchoverscore}), maximum width (Figure~\ref{fig:widthoverscore}), and maximum depth (Figure~\ref{fig:depthoverscore}) of these two groups of games in each decile.
Higher scoring games had, on average, instructions with significantly higher width and branching factor. 
But, games in the lower 50\% of scores increased their syntactic complexity more greatly over time. 
}

\eat{
\paragraph{Syntactic Change of Individual Users}\label{para:complexity_indiv}
We  analyzed how the syntactic complexity of individual leaders changed over their lifetime.
Figure~\ref{fig:IndivLifeCopmlexity} shows average branching factor, maximum width, and maximum depth averaged over individual lifetimes, measured by each player's $n$-th game.
}

\subsection{Reference Change}\label{sec:supp:refclass}
We divide environmental objects in the CerealBar game into six classes: road, foliage, building, water, rock, ice, and light class objects. We use regular expressions to automate if an utterance refers to a class of objects, defined by if it contains at least one of the class keywords in Table~\ref{tab:refclass}.

\begin{table}
\footnotesize
\begin{tabular}{lp{5cm}}
    \toprule
    Class & Keywords \\
    \midrule
    Road & \nlstring{road}, \nlstring{fork}, \nlstring{path}, \nlstring{intersect}, \nlstring{trail}, \nlstring{crossroad}, \nlstring{crosspath}, \nlstring{walkway} \\
    Foliage & \nlstring{palm}, \nlstring{flower}, \nlstring{tree}, \nlstring{shrub}, \nlstring{grass}, \nlstring{pine}, \nlstring{bush}, \nlstring{grove}, \nlstring{plant}, \nlstring{conif}, \nlstring{field}, \nlstring{foliag}, \nlstring{wasteland}, \nlstring{forest}, \nlstring{clearing}, \nlstring{patch}, \nlstring{lawn} \\
    Building & \nlstring{tower}, \nlstring{building}, \nlstring{house}, \nlstring{tent}, \nlstring{barn}, \nlstring{fort}, \nlstring{doghouse}, \nlstring{hut}, \nlstring{village}, \nlstring{cabin}, \nlstring{shack}, \nlstring{structure}, \nlstring{shed}, \nlstring{tower} \\
    Water & \nlstring{lake}, \nlstring{pond}, \nlstring{water}, \nlstring{sea}, \nlstring{river}, \nlstring{coast}, \nlstring{island}, \nlstring{shore} \\
    Rock & \nlstring{rock}, \nlstring{cliff}, \nlstring{boulder}, \nlstring{mountain}, \nlstring{hill}, \nlstring{log}, \nlstring{stone} \\
    Ice & \nlstring{glacier}, \nlstring{ice}, \nlstring{iceberg} \\
    Light & \nlstring{post}, \nlstring{lamp}, \nlstring{pole}, \nlstring{light} \\
    \bottomrule
\end{tabular}
\caption{Reference class keywords}
\label{tab:refclass}
\end{table}

%% file: supp_boxplots.tex
\begin{figure}[t]
\footnotesize
  \centering
  \begin{tikzpicture}\pgfplotsset{
      xtick={1,2,3,4,5,6,7,8,9,10},
      xmin=0, xmax=11
  }
  \begin{axis}
  [
    ymin=10, ymax=30,
    xlabel=Community Lifetime Decile,
    ylabel= \# Instructions in Interaction,
  legend pos=south east, legend style={nodes={scale=0.75, transform shape}}, width=6cm, height=4cm
  ]
  \addplot[mark=square*,plotcolor1]
    coordinates{
      (1,13.8)
      (2,16.3)
      (3,19.4)
      (4,21.7)
      (5,22.6)
      (6,23.5)
      (7,24.5)
      (8,25.1)
      (9,26.8)
      (10,26.7)
  };
  \end{axis}
  \end{tikzpicture} 
      \caption{Mean interaction length, measured by the number of instructions, in each decile. We include  incomplete instructions in these counts.}\label{fig:decile_interaction_length}
\end{figure}

\begin{figure}[t]
\footnotesize
  \centering
  \begin{tikzpicture}\pgfplotsset{
      xtick={1,2,3,4,5,6,7,8,9,10},
      xmin=0, xmax=11
  }
  \begin{axis}
  [
    ymin=7.5, ymax=9.5,
    xlabel=Community Lifetime Decile,
    ylabel=Follower Path Length,
  legend pos=south east, legend style={nodes={scale=0.75, transform shape}}, width=6cm, height=4cm
  ]
  \addplot[mark=square*,plotcolor1]
    coordinates{
    (1,7.957487922705314)
(2,8.312101910828025)
(3,8.44638069705094)
(4,8.130227001194744)
(5,8.760722347629796)
(6,8.611327040533038)
(7,8.456012493492972)
(8,8.915805785123966)
(9,8.712250712250713)
(10,8.824963432471964)
  }; %
  \end{axis}
  \end{tikzpicture} 
      \caption{Mean length of observed follower paths for complete instructions in each decile. We measure length in the number of steps recorded per instruction.}\label{fig:decile_path_length}
\end{figure}

%% file: cb-analysis.bbl
\begin{thebibliography}{19}
\expandafter\ifx\csname natexlab\endcsname\relax\def\natexlab#1{#1}\fi

\bibitem[{Clark and Marshall(1981)}]{Clark1981:definite-mutual-knowledge}
Herbert~H. Clark and Catherine~R. Marshall. 1981.
\newblock Definite knowledge and mutual knowledge.
\newblock \emph{Elements of discourse understanding}, pages 10--63.

\bibitem[{Clark and Wilkes{-}Gibbs(1986)}]{Clark1986-CLARAA-2}
Herbert~H. Clark and Deanna Wilkes{-}Gibbs. 1986.
\newblock \href {https://doi.org/10.1016/0010-0277(86)90010-7} {Referring as a
  collaborative process}.
\newblock \emph{Cognition}, 22(1):1--39.

\bibitem[{Djalali et~al.(2011)Djalali, Clausen, Lauer, Schultz, and
  Potts}]{Djalali11:cards-qud}
Alex Djalali, David Clausen, Sven Lauer, Karl Schultz, and Christopher Potts.
  2011.
\newblock Modeling expert effects and common ground using questions under
  discussion.
\newblock In \emph{AAAI Fall Symposium: Building Representations of Common
  Ground with Intelligent Agents}.

\bibitem[{Djalali et~al.(2012)Djalali, Lauer, and
  Potts}]{Djalali12:cards-preference}
Alex Djalali, Sven Lauer, and Christopher Potts. 2012.
\newblock Corpus evidence for preference-driven interpretation.
\newblock In \emph{Logic, Language and Meaning}.

\bibitem[{Gavran et~al.(2018)Gavran, Boldt, Darulova, and
  Majumdar}]{gavran2018precise}
Ivan Gavran, Brendon Boldt, Eva Darulova, and Rupak Majumdar. 2018.
\newblock \href {http://arxiv.org/abs/1803.02238} {Precise but natural
  specification for robot tasks}.

\bibitem[{Goodman and Frank(2016)}]{Goodman2016PragmaticLI}
Noah~D. Goodman and Michael~C. Frank. 2016.
\newblock Pragmatic language interpretation as probabilistic inference.
\newblock \emph{Trends in Cognitive Sciences}, 20:818--829.

\bibitem[{Haber et~al.(2019)Haber, Baumg{\"a}rtner, Takmaz, Gelderloos, Bruni,
  and Fern{\'a}ndez}]{Haber:19}
Janosch Haber, Tim Baumg{\"a}rtner, Ece Takmaz, Lieke Gelderloos, Elia Bruni,
  and Raquel Fern{\'a}ndez. 2019.
\newblock \href {https://www.aclweb.org/anthology/P19-1184} {The {P}hoto{B}ook
  dataset: Building common ground through visually-grounded dialogue}.
\newblock In \emph{Proceedings of the Annual Meeting of the Association for
  Computational Linguistics}.

\bibitem[{Hawkins et~al.(2020{\natexlab{a}})Hawkins, Frank, and
  Goodman}]{hawkins2020characterizing}
Robert X.~D. Hawkins, Michael~C. Frank, and Noah~D. Goodman.
  2020{\natexlab{a}}.
\newblock Characterizing the dynamics of learning in repeated reference games.
\newblock \emph{Cognitive science}, 44 6:e12845.

\bibitem[{Hawkins et~al.(2017)Hawkins, Frank, and
  Goodman}]{hawkins2017convention}
Robert X.~D. Hawkins, Mike Frank, and Noah~D. Goodman. 2017.
\newblock Convention-formation in iterated reference games.
\newblock In \emph{Cognitive Science}.

\bibitem[{Hawkins et~al.(2020{\natexlab{b}})Hawkins, Goodman, Goldberg, and
  Griffiths}]{Hawkins2020:partners-popluations}
Robert X.~D. Hawkins, Noah~D. Goodman, A.~Goldberg, and T.~Griffiths.
  2020{\natexlab{b}}.
\newblock Generalizing meanings from partners to populations: Hierarchical
  inference supports convention formation on networks.
\newblock In \emph{Proceedings of the Annual Conference of the Cognitive
  Science Society}.

\bibitem[{He et~al.(2017)He, Balakrishnan, Eric, and Liang}]{He:17dialogue}
He~He, Anusha Balakrishnan, Mihail Eric, and Percy Liang. 2017.
\newblock \href {https://doi.org/10.18653/v1/P17-1162} {Learning symmetric
  collaborative dialogue agents with dynamic knowledge graph embeddings}.
\newblock In \emph{Proceedings of the Annual Meeting of the Association for
  Computational Linguistics}.

\bibitem[{Honnibal and Montani(2017)}]{spacy2}
Matthew Honnibal and Ines Montani. 2017.
\newblock {spaCy 2}: Natural language understanding with {B}loom embeddings,
  convolutional neural networks and incremental parsing.
\newblock To appear.

\bibitem[{Krauss and Weinheimer(1964)}]{Krauss1964ChangesIR}
Robert~M. Krauss and Sidney Weinheimer. 1964.
\newblock Changes in reference phrases as a function of frequency of usage in
  social interaction: a preliminary study.
\newblock \emph{Psychonomic Science}, 1:113--114.

\bibitem[{Monroe et~al.(2017)Monroe, Hawkins, Goodman, and
  Potts}]{monroe2017colors}
Will Monroe, Robert~X.D. Hawkins, Noah~D. Goodman, and Christopher Potts. 2017.
\newblock \href {https://doi.org/10.1162/tacl_a_00064} {Colors in context: A
  pragmatic neural model for grounded language understanding}.
\newblock \emph{Transactions of the Association for Computational Linguistics},
  5:325--338.

\bibitem[{Potts(2012)}]{Potts:12}
Christopher Potts. 2012.
\newblock Goal-driven answers in the {C}ards dialogue corpus.
\newblock In \emph{Proceedings of the West Coast Conference on Formal
  Linguistics}, pages 1--20.

\bibitem[{Suhr et~al.(2019)Suhr, Yan, Schluger, Yu, Khader, Mouallem, Zhang,
  and Artzi}]{Suhr2019:cerealbar}
Alane Suhr, Claudia Yan, Jack Schluger, Stanley Yu, Hadi Khader, Marwa
  Mouallem, Iris Zhang, and Yoav Artzi. 2019.
\newblock \href {https://doi.org/10.18653/v1/D19-1218} {Executing instructions
  in situated collaborative interactions}.
\newblock In \emph{Proceedings of the Conference on Empirical Methods in
  Natural Language Processing}.

\bibitem[{Udagawa and Aizawa(2019)}]{Udagawa:19}
Takuma Udagawa and Akiko Aizawa. 2019.
\newblock A natural language corpus of common grounding under continuous and
  partially-observable context.
\newblock In \emph{Proceedings of the Conference on Artificial Intelligence}.

\bibitem[{Wang et~al.(2017)Wang, Ginn, Liang, and
  Manning}]{wang-etal-2017-naturalizing}
Sida~I. Wang, Samuel Ginn, Percy Liang, and Christopher~D. Manning. 2017.
\newblock \href {https://doi.org/10.18653/v1/P17-1086} {Naturalizing a
  programming language via interactive learning}.
\newblock In \emph{Proceedings of the 55th Annual Meeting of the Association
  for Computational Linguistics (Volume 1: Long Papers)}, pages 929--938,
  Vancouver, Canada. Association for Computational Linguistics.

\bibitem[{Xu and Reitter(2016)}]{xu-reitter-2016-convergence}
Yang Xu and David Reitter. 2016.
\newblock \href {https://doi.org/10.18653/v1/P16-2072} {Convergence of
  syntactic complexity in conversation}.
\newblock In \emph{Proceedings of the Annual Meeting of the Association for
  Computational Linguistics}.

\end{thebibliography}
